\documentclass[runningheads]{llncs}
\usepackage[numbers, compress]{natbib}
\usepackage[usenames,dvipsnames]{pstricks}
\usepackage[utf8]{inputenc} 
\usepackage[T1]{fontenc}    
\usepackage{lmodern}
\usepackage{hyperref}       
\usepackage{url}            
\usepackage{multirow,booktabs,longtable}       
\usepackage{amsfonts}       
\usepackage{nicefrac}       
\usepackage{microtype}      
\usepackage[pdftex]{graphicx}
\usepackage[toc,page]{appendix}
\usepackage{tikz}
\usetikzlibrary{snakes}
\usepackage{subcaption}
\usepackage[percent]{overpic}
\usepackage{caption}
\usepackage{array}
\usepackage{etex}

\usepackage[fleqn]{amsmath}
\newcolumntype{P}[1]{>{\centering\arraybackslash}p{#1}}

\usepackage{float}
\usepackage{aliascnt}
\newaliascnt{eqfloat}{equation}
\newfloat{eqfloat}{h}{eqflts}
\floatname{eqfloat}{Equation}
\newcommand*{\ORGeqfloat}{}
\let\ORGeqfloat\eqfloat
\def\eqfloat{%
	\let\ORIGINALcaption\caption
	\def\caption{%
		\addtocounter{equation}{-1}%
		\ORIGINALcaption
	}%
	\ORGeqfloat
}

\title{Longitudinal detection of radiological abnormalities with time-modulated LSTM}

\author{
	Ruggiero Santeramo \inst{1,2}\and Samuel Withey\inst{3} \and Giovanni Montana\inst{1,2}
}
\institute{Department of Biomedical Engineering, King's College London 	\email{\{ruggiero.santeramo,giovanni.montana\}@kcl.ac.uk}
	\and
	WMG, University of Warwick
	\email{\{ruggiero.santeramo,G.montana\}@warwick.ac.uk}
	\and
	Department of Radiology, Guy’s \& St Thomas’ NHS Foundation Trust}
\begin{document}

\maketitle

\begin{abstract}
Convolutional neural networks (CNNs) have been successfully employed in recent years for the detection of radiological abnormalities in medical images such as plain x-rays. To date, most studies use CNNs on individual examinations in isolation and discard previously available clinical information. In this study we set out to explore whether Long-Short-Term-Memory networks (LSTMs) can be used to improve classification performance when modelling the entire sequence of radiographs that may be available for a given patient, including their reports. A limitation of traditional LSTMs, though, is that they implicitly assume equally-spaced observations, whereas the radiological exams are event-based, and therefore irregularly sampled. Using both a simulated dataset and a large-scale chest x-ray dataset, we demonstrate that a simple modification of the LSTM architecture, which explicitly takes into account the time lag between consecutive observations, can boost classification performance. Our empirical results demonstrate improved detection of commonly reported abnormalities on chest x-rays such as cardiomegaly, consolidation, pleural effusion and hiatus hernia.

\end{abstract}

\section{Introduction} \label{introduction}

Deep learning approaches have exhibited impressive performance in medical imaging applications in recent years \cite{chest-ex,deep_ref2,deep_ref3}. For instance, convolutional neural networks (CNNs) have had some success in detecting and classifying radiological abnormalities on chest x-rays, a particularly complex task \cite{chest-ex,chest1,chest-ex2, chest-survey}. The majority of these studies have been designed for cross-sectional analyses, viewing a single image in isolation, and discard the fact that a patient may have had previous medical imaging examinations for which the radiological reports are also available. It is standard practice for radiologists to take clinical history into account to add context to their report by using comparison to previous imaging. Some abnormalities will be long-standing, but others may change over time, with varying clinical relevance. Often in elderly patients or those with a history of smoking, the baseline x-ray appearances, i.e. when that patient is ‘well’, can still be abnormal. If individual films are viewed in isolation, it can be challenging to tell with certainty if there are acute findings. If previous imaging is available, it is possible to determine if there has been interval change, for example, acute consolidation (indicating infection). As with humans, it is expected that a neural network can learn from previous patient-specific information, in this case all prior chest radiographs for that patient and their corresponding reports.

\begin{figure}[h]
	\centering
	\captionsetup{width=0.8\textwidth}
	\fbox{\rule[0cm]{0cm}{0cm} 
	\centering
\begin{minipage}[H]{1\linewidth}
\centering
\vspace{1ex}

\vspace{1ex}

\begin{minipage}[H]{0.90\linewidth}

\begin{subfigure}[ht]{0.255\linewidth}
	\centering
	
	\begin{overpic}[width=1\linewidth]{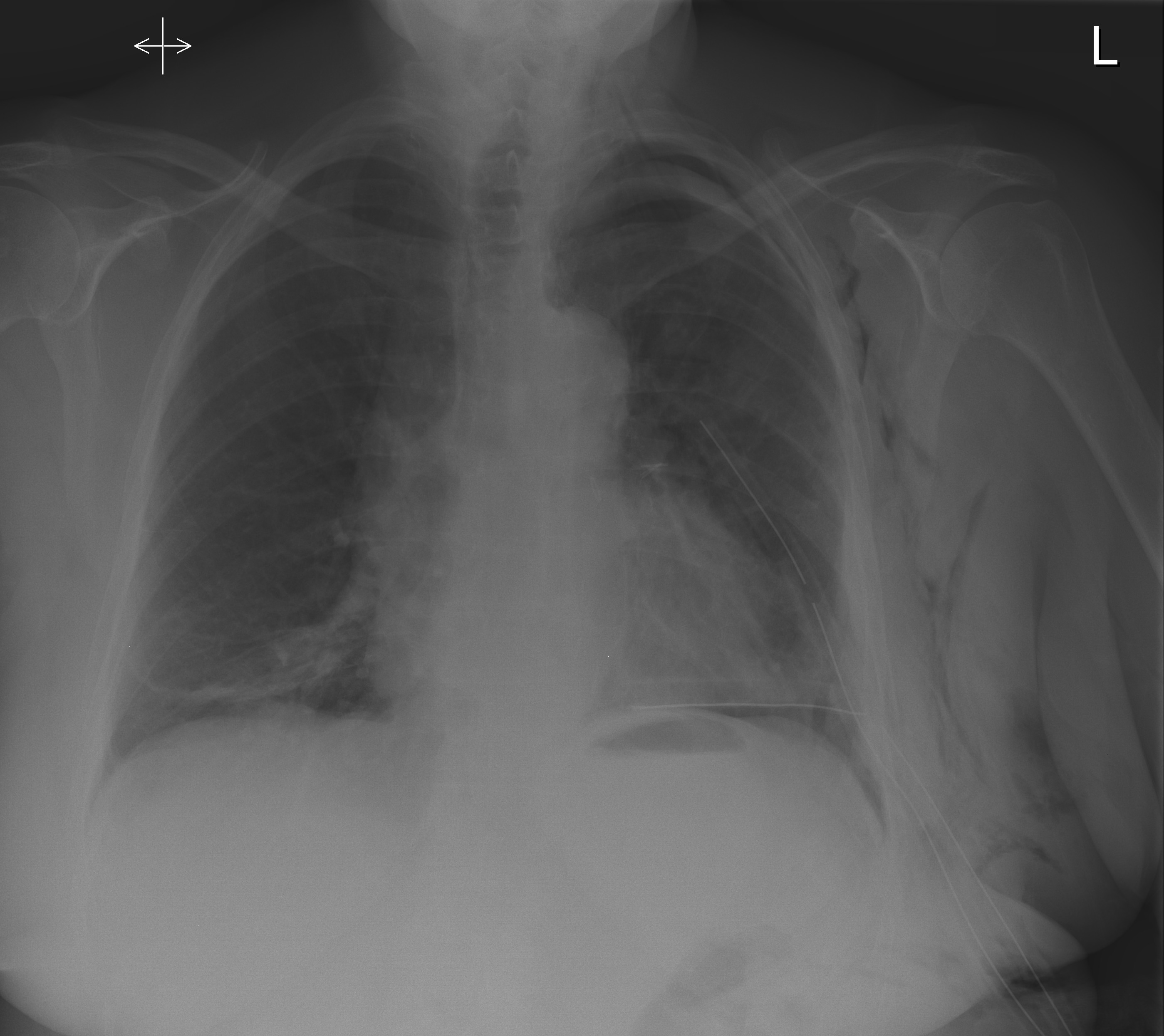}
		\put(2.5,4.5){\color{white}\small$X_i^{t_0}$}
	\end{overpic}
	\footnotesize{\tiny\label{fig:timeline1_A} medical device, pneumothorax, emphysema, atelectasis.}
	\vspace{0ex}
\end{subfigure}
\begin{subfigure}[ht]{0.23\linewidth}
	\centering
	\begin{overpic}[width=1\linewidth]{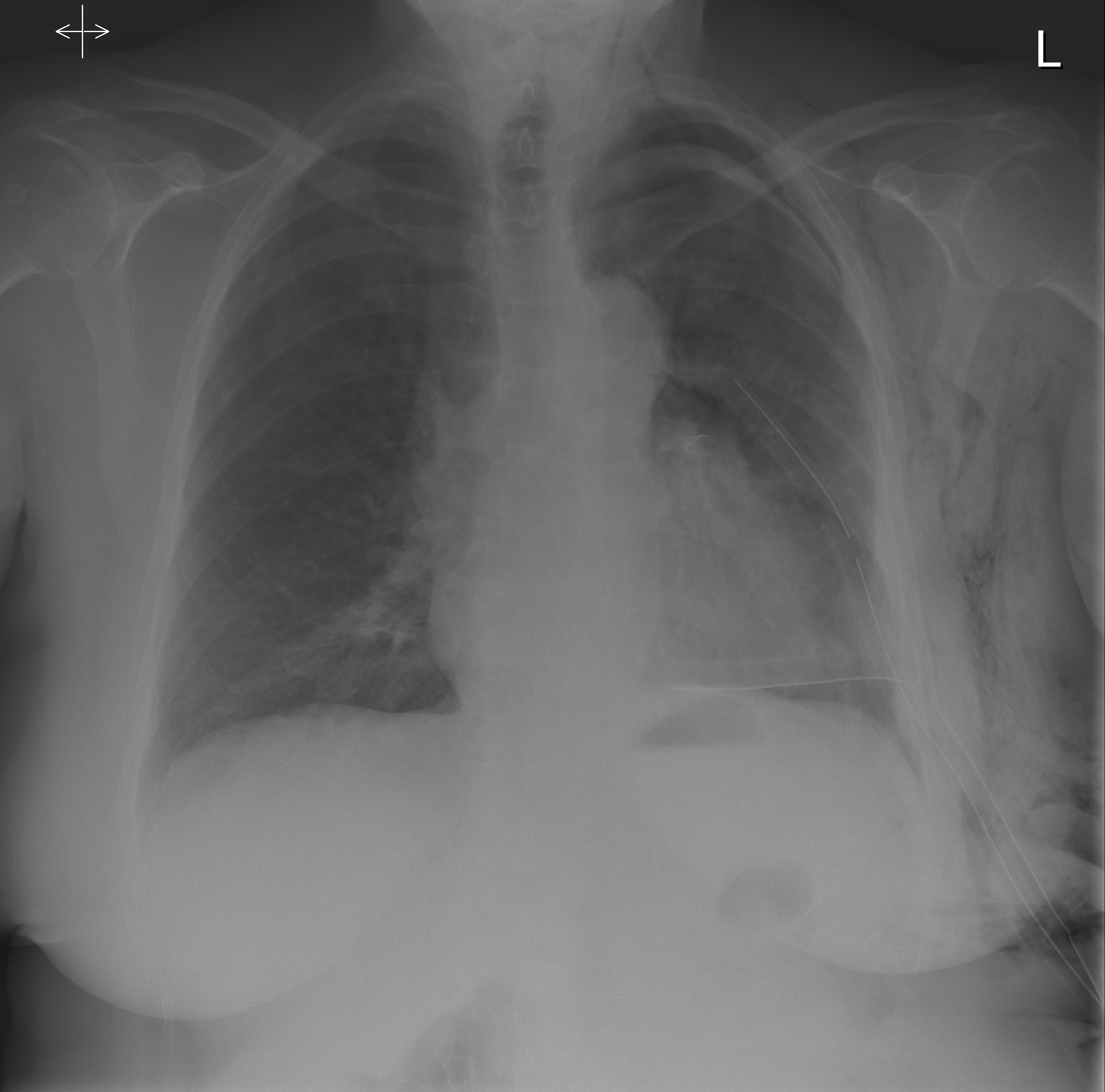}
	\put(2.5,4.5){\color{white}\small$X_i^{t_2}$}
\end{overpic}
	\footnotesize{\tiny\label{fig:timeline1_B} medical device, emphysema and pneumothorax.}
	\vspace{0ex}
\end{subfigure}
\begin{subfigure}[ht]{0.23\linewidth}
	\centering
	\begin{overpic}[width=1\linewidth]{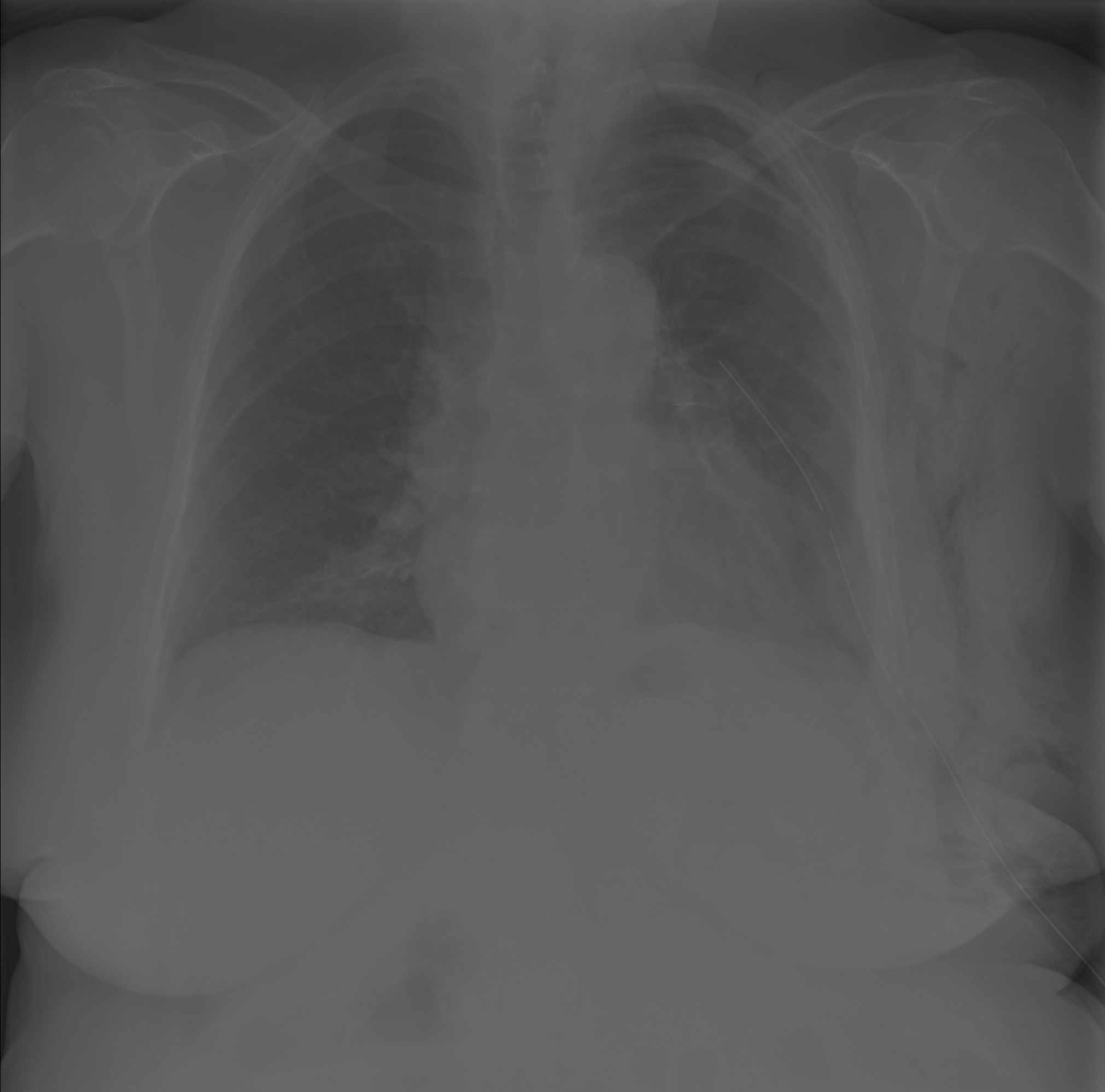}
	\put(2.5,4.5){\color{white}\small$X_i^{t_3}$}
	
\end{overpic}
	\footnotesize{\tiny\label{fig:timeline1_C} medical device, emphysema and pneumothorax.}
	\vspace{0ex}
\end{subfigure}
\begin{subfigure}[ht]{0.255\linewidth}
	\centering
	\begin{overpic}[width=1\linewidth]{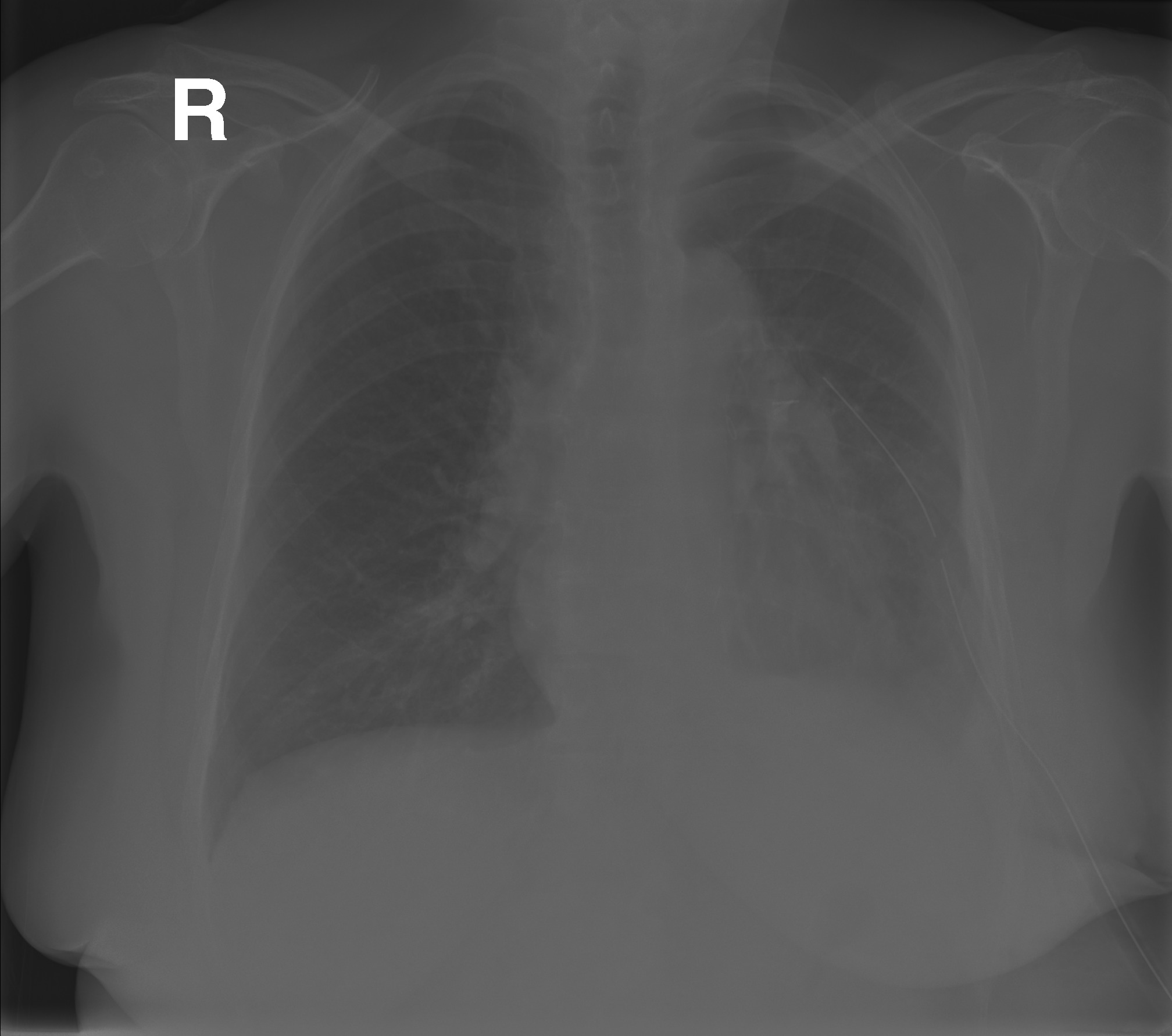}
	\put(2.5,4.5){\color{white}\small$X_i^{t_{15}}$}
\end{overpic}
	\footnotesize{\tiny\label{fig:timeline1_D} pneumothorax, emphysema and medical device.}
	\vspace{0ex}
\end{subfigure}

\begin{minipage}{1\linewidth}
\centering
\begin{tikzpicture}[snake=zigzag, line before snake = 2mm, line after snake = 2mm, scale=0.6]
\centering

\draw (0,0) -- (15,0);
\draw[->,>=stealth] (15,0) -- node[below left] {} (16,0);

\foreach \x in {0,2,3,15}
\draw (\x cm,3pt) -- (\x cm,-3pt);

\draw (0,0) node[above=3pt] {$ X_i^{t_0} $};
\draw (2,0) node[above=3pt] {$ X_i^{t_2} $};
\draw (3,0) node[above=3pt] {$ X_i^{t_3} $};
\draw (15,0) node[above=3pt] {$ X_i^{t_{15}} $};

\end{tikzpicture}
\end{minipage}
\vspace{0 ex}
\end{minipage}

\end{minipage}
		\rule[0cm]{-.12cm}{0cm}}
	\caption{\protect\label{fig:timeline1}Example of longitudinal x-rays for a  given patient.}
	\vspace{-0.5cm}
\end{figure}
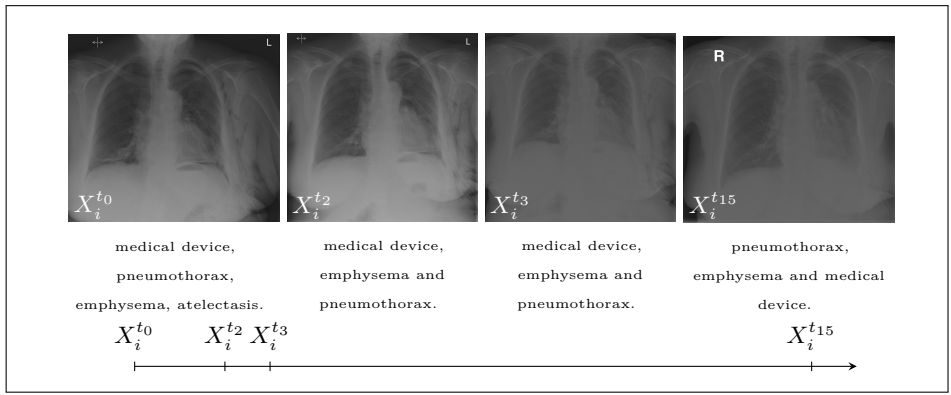

The motivation for this work is to assess the  potential of recurrent neural networks (RNNs) for the real-time detection of radiological abnormalities when modelling the entire series of past exams that are available for any given patient. In particular, we set out to explore the performance of Long Short-Term Memory (LSTM) networks \cite{lstm2, lstm}, which have lately become the method of choice in sequential modelling, especially when used in combination with CNNs for visual feature extraction \cite{CNN-LSTM,lstm-cnn-2}. The technical challenge faced in our context is that sequential medical exams are event-based observations. As such, they are collected at times of clinical need, i.e. they are not equally spaced, and the number of historical exams available for each patient can vary greatly. Fig. \ref{fig:timeline1} shows four longitudinal chest x-rays acquired on the same patient over a certain period of time. This figure also illustrates other challenges faced when modelling this type of longitudinal data: the images may be aquired using different x-ray devices (resulting in different image quality, i.e. resolution, brightness, etc.), there may be differences in patient positioning (i.e. supine, erect, rotated, degree of inspiration), differences in projection (postero-anterior and antero-posterior), and not all images are equally centred (i.e. there can be rotations, translations, etc.). 

As LSTMs are typically applied on regularly-sampled data \cite{lstm-task1,lstm-task2, lstm-task3}, they are ill-suited to work with irregular time gaps between consecutive observations, as previously noted \cite{phase-lstm, t-lstm-reviewer}. This is a particularly important limitation in our context as certain radiological abnormalities tend to be observed for longer periods of time whereas others are short-lived. In this article we demonstrate that an architecture combining a CNN with a simple modification of the standard LSTM is able to handle irregularly-sampled data and learn the temporal dynamics of certain visual features resulting in improved pattern detection. Using both simulated and real x-ray datasets, we demonstrate that this capability yields improved image classification performance over an LSTM baseline.

\section{Motivating dataset and problem formulation} \label{dataset}

The dataset used in this study was collected from the historical archives of the PACS (Picture Archiving and Communication System) at Guy's and St. Thomas' NHS Foundation Trust, in London, during the period from January 2005 to March 2016. The dataset has been previously used for the detection of lung nodules \cite{epesce} and for multi-label metric learning \cite{mauro}. It consists of $745\,480$ chest radiographs representative of an adult population and acquired using 40 different x-ray systems. Each associated radiological report was parsed using a natural language processing system for the automated extraction of radiological labels \cite{epesce, nlp}. For this study, we extracted a subset of $80\,737$ patients having a history of at least two exams, which resulted in $337\,575$ images (with $232\,610$ used for training and $104\,965$ for testing). Each image was scaled to a standard format of $299 \times 299$ pixels. The resulting dataset has an average of $4.18$ examinations per patient with an average of $180.29$ days between consecutive exams per patient. 

In what follows, each individual sequence of longitudinal chest x-rays along with its associated vector of radiological labels is denoted as $\{X_{i}^t, l_{i}^t\}$, where $i=1,\ldots,N$ is the patient index and $t=1, \ldots,T_i$ is the time index. Typical chest x-ray datasets are characterised by relatively few examinations per patient (e.g. $T_i$ is around 4-5) and highly-irregular sampling rates. Our task is to predict the vector of image labels $l_{i}^{T_i}$ given the entire history of exams up to time $T_i-1$ plus the current image, i.e. $X_i^{T_i}$. 

\section{Time-modulated LSTM} \label{tmlstm}

LSTMs are a particular type of RNNs able to classify, process and predict time series \cite{lstm, lstm2}. The internal state of an LSTM (a.k.a. the cell state or memory) gives the architecture its ability to 'remember'.  A standard LSTM contains memory blocks, and blocks contain memory cells. A typical memory block is made of three main components: an input gate controlling the flow of input activations into the memory cell, an output gate controlling the output flow of cell activations, and a forget gate for scaling the internal state of the cell. The forget gate modulates how much information is used from the internal state of the previous time-step. However, standard LSTMs are ill-suited for our task where the time between consecutive exams is variable, because they have no mechanism for explicitly modelling the arrival time of each observation. In fact, it has been shown that LSTMs, and more generally RNNs, underperform with irregularly sampled data or time series with missing values \cite{phase-lstm, che2018recurrent}. Previous attempts to adapt LSTMs for use with irregularly sampled datapoints have mostly focused on speeding up the converge of the algorithm in settings with high-resolution sampled data \cite{phase-lstm} or to discount short-term memory \cite{t-lstm-reviewer}.

\begin{figure}[t]		
		\centering
		\fbox{\rule[0cm]{0cm}{0cm}
			\rule[0cm]{0cm}{0cm}
			\includegraphics[width=1.\linewidth]{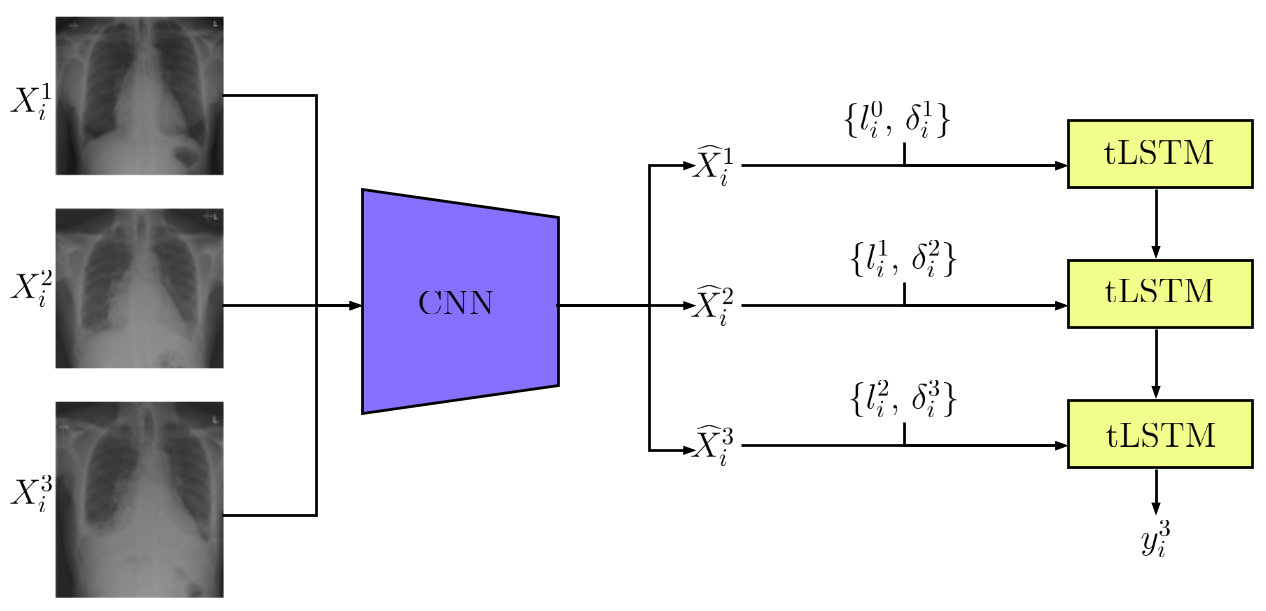}
			\rule[0cm]{0cm}{0cm}}
		\caption{\label{fig:LSTM} An overview of the proposed architecture for image label prediction leveraging all historical exams.} 
		\vspace{-0.5cm}
\end{figure}

To address these issues, we introduce two simple modifications of the standard LSTM architecture, called time-modulated LSTM (tLSTM), both making explicit use of the time indexes associated to the inputs. In the proposed architecture, all the images for a given patient are initially processed by a CNN architecture, which extracts a set of imaging features, denoted by $\widehat{X}_i^t$, at each time step. The LSTM takes as inputs $l_i^{t-1}$, i.e. the radiological labels describing the images acquired at the previous time-step, the current image features, $\widehat{X}_i^t$, and the time lapse between $X_i^{t-1}$ and $X_i^{t}$, which we denote as $\delta_i^t$. For the last image in the sequence, the LSTM predicts the image labels, $l_i^t$, called $y_i^t$. Figure \ref{fig:LSTM} provides a high-level overview of this model and the equations below define the tLSTM unit:
\begin{equation} \label{eqn:lstm-eq}
\begin{aligned}
f_t & = \sigma(W_{fl}*l^{t-1} + W_{fx}*\widehat{X}^t + W_{fj}*\delta^t + b_f) ,\\
i_t & = \sigma(W_{il}*l^{t-1} + W_{ix}*\widehat{X}^t + W_{ij}*\delta^t + b_i) ,\\
o_t & = \sigma(W_{ol}*l^{t-1} + W_{ox}*\widehat{X}^t + W_{oj}*\delta^t + b_o) ,\\
c_t & = \tanh(W_{cl}*l^{t-1} + W_{cx}*\widehat{X}^t + W_{cj}*\delta^t + b_c) ,\\
h_t & = f_t * h_{t-1} + i_t * c_t ,\\
y^t & = o_t * \tanh(h_t) 
\end{aligned}
\end{equation}
Here, $h_t$ defines the internal state at time-step $t$, while $f_t$, $i_t$ and $o_t$ refer to the forget, input and output gates at time-step $t$, respectively. These are all computed as linear combinations of the vectors $l^{t-1}, \widehat{X^t}$ and the scalar $\delta^t$, and then transformed by a sigmoid function, $\sigma(\cdot)$. The matrices denoted by $W$ contain learnable weights indexed by two letters (e.g. $W_{fl}$ contains the weights of the forget gate $f$ for labels $l$, and so on). At time $t = 1$, we initialise $l_i^{t-1} = <0\dots0>$ (an array of zeros) and $\delta_i^t=0$. The time lapses, $\delta_i^t$, linearly modulate the information inside the internal cell state as well as the output, forget and input gates. 

A different variation of the previous model (tLSTMv2) uses the time lapse only to modulate the internal state, $h_t$. In this case, each $\delta_i^t$ actively contributes to updating $h_t$ directly and, implicitly, to estimating the label vector $y^t$, i.e. 
\begin{equation}
	\label{eqn:lstm-eq2}
	\begin{aligned}
	h_t &= f_t * h_{t-1} + i_t * c_t + W_{tj}*\delta^t\\
	y^t &= o_t * \tanh(h_t) .
	\end{aligned}
	\end{equation}
The form of the other updating equations, i.e. $f_g, i_t, o_t$ and $c_t$, is similar to those in Eq. \eqref{eqn:lstm-eq}, without the $Ws \times \delta^t$ elements. 
	
\section{Simulated data} \label{simulated-data}

In order to better assess the potential advantages introduced by the time-modulated LSTM in settings where observations are event-driven and the underlying patterns to be detected are time-varying, we generated simulated data as an alternative to the the real chest x-ray dataset of Section \ref{dataset}. Simulating images enables us to precisely control the sampling frequency at which the relevant visual patterns appear and disappear over time as well as the signal to noise ratio. For this study, we simulated a population of image sequences of varying lengths. Within a sequence, each image consisted of a noisy background image containing one or more randomly placed digits drawn from the set  $\{0, 3, 6, 8, 9\}$. We simulated three  kinds of patterns inspired by the radiological patterns seen in real medical images: (i) {\it rare patterns} consisting of digits appearing with low probability; (ii) {\it common patterns} consisting of rapidly appearing and resolving digits; (iii) {\it persistent labels}, consisting of digits observed for extended periods of time. In analogy to medical images, each digit in our simulation represents a radiological abnormality to be detected, hence multiple (and possibly overlapping) digits are allowed to coexist within an image. The time lapse $\delta^t$ was modelled as a uniform random variable taking value in the interval $[1,10]$. An example of simulated images can be found in the Supplementary Material.

\section{Experimental results} \label{results}

In our experiments with the real x-ray dataset, the CNN component in our architecture conists of a pre-trained Inception v3 \cite{inception-v3} without the classification layer. The imaging features $\hat{X}_i^t$ (an array $2048$ elements) from the CNN are as used as inputs for the LSTM component along with the image labels. We considered four possible radiological labels: cardiomegaly, consolidation, pleural effusion and hiatus hernia. The performance of the time-modulated LSTM models is assessed by the PPV (Positive Predictive Value) and NPV (Negative Predictive Value) along with F-score, i.e the harmonic mean of precision and recall.  

\begin{longtable}{r P{1.7cm} P{1.7cm} P{1.7cm} P{1.7cm} P{1.7cm}}
	\caption{Results on real data\textsuperscript{*}}\\*
	\label{tab:results-real-table}
	\centering
	\begin{tabular}{r P{1.7cm} P{1.7cm} P{1.7cm} P{1.7cm} P{1.7cm}}
		\toprule
		\multirow{4}{*}{\textbf{Inception v3}} & \multicolumn{5}{c}{\textbf{Labels}} \\
		\midrule
		& \textbf{cardio.} & \textbf{consol.} & \textbf{pleu. eff.} & \textbf{hernia} & \textbf{avg.}\\
		\cmidrule(lr){2-6}
		\textbf{PPV} & \textit{$0.5477$} &\textit{$0.4111$}&
		\textit{$0.6149$}& \textit{$0.5204$} & \textit{$0.5235$} \\
		\textbf{NPV} & \textit{$0.9565$} &\textit{$0.9002$}&
		\textit{$0.9106$}& \textit{$0.9958$}& \textit{$0.9407$} \\
		\textbf{F-measure} & \textit{$0.6143$} &\textit{$0.5151$}&
		\textit{$0.6575$}& \textit{$0.5193$}& \textit{$0.5765$} \\
		\midrule
		\textbf{LSTM} &&&&&\\
		\cmidrule(lr){2-6}
		\textbf{PPV} & \textit{$0.6914$} &\textit{$0.5841$}&
		\textit{$0.7105$}& \textit{$0.5369$}&\textit{$0.6307$} \\
		\textbf{NPV} & \textit{$0.9406$} &\textit{$0.8440$}&
		\textit{$0.8895$}& \textit{$0.9969$}&\textit{$0.9177$} \\
		\textbf{F-measure} & \textit{$0.6199$} &\textit{$0.4337$}&
		\textit{$0.6531$}& \textit{$0.5755$}&\textit{$0.5705$} \\
		\midrule
		\textbf{tLSTMv1} &&&&&\\
		\cmidrule(lr){2-6}
		\textbf{PPV} & \textit{$0.5929$} &\textit{$0.4831$}&
		\textit{$0.6358$}& \textit{$0,5821$}&\textit{$0.5734$} \\
		\textbf{NPV} & \textit{$0.9565$} &\textit{$0.9000$}&
		\textit{$0.9251$}& \textit{$0.9968$}&\textit{$0.9445$} \\
		\textbf{F-measure} & \textit{$0.6399$} &\textit{$0.5552$}&
		\textit{$0.6891$}& \textit{$0.5932$}&\textit{$0.6193$} \\
		
		\midrule
		\textbf{tLSTMv2} &&&&&\\
		\cmidrule(lr){2-6}
		\textbf{PPV} & \textit{$0.5980$} &\textit{$0.4876$}&
		\textit{$0.6350$}& \textit{$0.5461$}&\textit{$0.5667$} \\
		\textbf{NPV} & \textit{$0.9572$} &\textit{$0.8931$}&
		\textit{$0.9120$}& \textit{$0.9968$}&\textit{$0.9397$} \\		
		\textbf{F-measure} & \textit{$0.6447$} &\textit{$0.5479$}&
		\textit{$0.6696$}& \textit{$0.5704$}&\textit{$0.6081$} \\
		
		\cmidrule(lr){1-6}
		\multicolumn{6}{l}{}
		\begin{minipage}{10cm}
			\textsuperscript{*}\footnotesize{Classification performance (PPV, NPP and F-measure) of a baseline classifier (Inception v3) using only a single image as inpur and three LSTM architectures using the full sequence of longitudinal observations. tLSTMv1 and tLSTMv2 are the proposed time-modulated LSTM architectures that explitely model time lapses.}
		\end{minipage}
		\\
		\bottomrule
		
	\end{tabular}
	
\end{longtable}

We compared the performance of four models: the baseline CNN classifier (Inceptionv3) that only uses each  current image to predict the labels, but does not exploit the historical exams for a given patient, and three variations of the architecture illustrated in Figure \ref{fig:LSTM}: one using the standard LSTM and the two versions of time-modulated LSTM model introduced in Section \ref{tmlstm}. Both tLSTM versions introduced noticeable performance improvements; see Table \ref{tab:results-real-table}. In particular, tLSTMv1 yields an increase of $\sim7$\% in F-measure over the baseline and $\sim8$\% over a standard LSTM. Moreover, tLSTMv1 achieves a $\sim9$\% improvement in PPV over the baseline. Overall, tLSTM achieves improved performance over the standard LSTM due to its ability to handle irregularly sampled data.

For the simulated dataset, we used a pre-trained AlexNet \cite{alexnet} as feature extractor in combination with three versions of the LSTM for modelling sequences of images. 
A full table with results can be found in the Supplementary Material. We purposely introduced a sufficiently high level of noise in the visual patterns so as to make the classification problem with individual images particularly difficult; accordingly, the single-image classifier did not achieve acceptable classification results. Likewise, the architecture using a standard LSTM did not introduce significant improvements due to the irregularly sampled observations. On the other hand, larger classification improvements were achieved using the time-modulated LSTM units as those were able to decode the sequential patterns by explicitly taking into account the time gaps between consecutive observations. 

\section{Conclusions} \label{conslusions}

Our experimental results suggest that the modified LSTM architectures, combined with CNNs, are suitable for modelling sequences of event-based imaging observations. By explicitly modelling the individual time lapses between consecutive events, these architectures are able to better capture the evolution of visual patterns over time, which has a boosting effect on the classification performance. The full potential of these models is best demonstrated using simulated datasets whereby we have control over the exact nature of the temporal patterns and the image labels are perfectly known. In real radiological datasets, there are often errors in some of the image labels due to typographical errors, interpretive errors, ambiguous language and, in some cases, long-standing findings not being mentioned. This can cause problems both in CNN training and testing. Despite these challenges, we have demonstrated that improved classification results can also be achieved by the time-modulated LSTM components on a large chest x-ray dataset. Thus we empirically proved that a patient's imaging history can be used to improve automated radiological reporting. In future work, we plan more extensive testing of a system trained end-to-end on a much larger number of radiological classes. The code with the networks used for our experiment can be found online: {\tt https://github.com/WMGDataScience/tLSTM}.

{
\renewcommand\bibsection{\section{\refname}}
\bibliographystyle{plainnat}
\bibliography{bib}
}
\section{Appendix: Artificial data simulation}
\begin{figure}[h]
	\centering
	\captionsetup{width=0.8\textwidth}
	\fbox{\rule[0cm]{0cm}{0cm} 
		\centering
		\begin{minipage}[H]{1\linewidth}
			\centering
			\vspace{1ex}

			\begin{minipage}[ht]{0.90\linewidth}

				\begin{subfigure}[ht]{0.245\linewidth}
					\centering
					
					\begin{overpic}[width=1\linewidth]{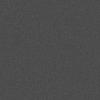}
						\put(2.5,2.5){\color{white}\small$X_1$}
					\end{overpic}
					\footnotesize{\tiny\label{fig:timeline2_A} None.}
					\vspace{0ex}
				\end{subfigure}
				\begin{subfigure}[ht]{0.245\linewidth}
					\centering
					\begin{overpic}[width=1\linewidth]{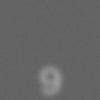}
						\put(2.5,2.5){\color{white}\small$X_2$}
					\end{overpic}
					\footnotesize{\tiny\label{fig:timeline2_B} '9'.}
					\vspace{0ex}
				\end{subfigure}
				\begin{subfigure}[ht]{0.245\linewidth}
					\centering
					\begin{overpic}[width=1\linewidth]{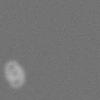}
						\put(2.5,2.5){\color{white}\small$X_3$}
						
					\end{overpic}
					\footnotesize{\tiny\label{fig:timeline2_C} '0' and '9'.}
					\vspace{0ex}
				\end{subfigure}
				\begin{subfigure}[ht]{0.245\linewidth}
					\centering
					\begin{overpic}[width=1\linewidth]{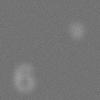}
						\put(2.5,2.5){\color{white}\small$X_4$}
					\end{overpic}
					\footnotesize{\tiny\label{fig:timeline2_D} '6' and '9'.}
					\vspace{0ex}
				\end{subfigure}
				\begin{minipage}{1\linewidth}
					\centering
					\begin{tikzpicture}[snake=zigzag, line before snake = 2mm, line after snake = 2mm, scale=1.3]
					\centering
					
					\draw (0,0) -- (8.6,0);
					\draw[->,>=stealth] (8.6,0) -- node[below left] {} (8.8,0);
					
					\foreach \x in {2.6,4.0,7.0,8.5}
					\draw (\x cm,3pt) -- (\x cm,-3pt);

					\draw (2.6,0) node[above=3pt] {$ X_i^{t_{26}} $};
					\draw (4.0,0) node[above=3pt] {$ X_i^{t_{40}} $};
					\draw (7.0,0) node[above=3pt] {$ X_i^{t_{70}} $};
					\draw (8.5,0) node[above=3pt] {$ X_i^{t_{85}} $};
					\end{tikzpicture}
				\end{minipage}
			\end{minipage}
		\end{minipage}
		\rule[0cm]{-.12cm}{0cm}}	
	\caption{\protect\label{fig:timeline2}Example of a simulated sequence of images with labels.}	
\end{figure}
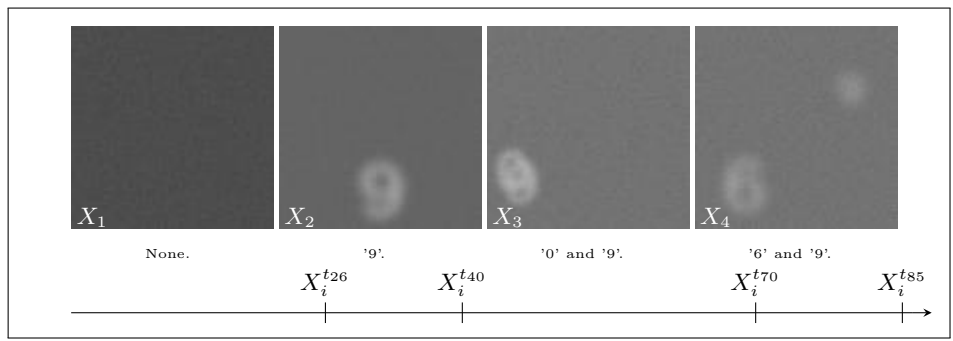


In this section we describe the data simulation procedure. Each data point consists of a variable number of simulated images. The length of each sequence of images is allowed to vary from a minimum of $10$ to a maximum of $100$, with an average of $20$ images per each sequence. All the simulated images consist of a prefixed grey background, some random noise (a Gaussian blur) applied to this background and simulated digits from a set ${0, 6, 8, 3, 9}$. 

At each time step, $t$, we draw a random integer $\delta_t \in [1,10]$ to represent the time elapsed between two consecutive images, $S_t$ and $S_{t+1}$. Initially, at state $S_0$, the image contains no digits. The digits allowed to be sampled at $S_{t+1}$ depend on the current state, $S_{t}$, and the particular value of $\delta_t$. Table \ref{tab:transition-table} defines all the digits allowed to be seen at $S_{t+1}$ as a function of $\delta_t$ and $S_{t}$. Each one of the allowed digits is then sampled with fixed probability. The digit is placed at a random location and its rotation angle is also randomly chosen.

According to this procedure, different digits behave differently, the digit "9" is independent of other labels, whilst all the others are dependent of each other. Some labels (e.g. "6" or "9") can persist over longer periods of time; some digits are rare, i.e. have a low probability of appearing (e.g. "0") while others are more frequent, i.e. have higher probabilities (e.g. "3"). These scenarios somewhat mimic rare and common abnormalities. Figure \ref{fig:timeline2} provides an example of a typical sequence, in this case with only 4 simulated images, each one having one or two digits.

Using this procedure, we simulated independent training and testing datasets used in our work to test and compare our models with the standard LSTM and the AlexNet as explained in the paper. The empirical results obtained on the test dataset can be found in Table \ref{tab:results-sim-table}.


\begin{table}[ht]
	\caption{State transition table\textsuperscript{*}} \label{tab:transition-table}
	\centering
	\begin{tabular}{r*{6}{c}}
		\toprule
		\multirow{4}{*}{\textbf{Current state ($S_t$)}} & \multicolumn{6}{c}{\textbf{Next state ($S_{t+1}$)}} \\
		\cmidrule(lr){2-7}
		& \textbf{0} & \textbf{6} & \textbf{8} & \textbf{3} & \textbf{9} & \textbf{null} \\
		\midrule
		\textbf{0} & \textit{7:10} &\textit{5}&
		\textit{3,7}& \textit{1,2,7} & 			 
		\textit{-}& \textit{1:6} \\
		\textbf{6} & \textit{1,2} &\textit{1:3,5:9}&
		\textit{3,6}& \textit{5} & 			 
		\textit{-}&\textit{4,10} \\
		\textbf{8} & \textit{1} &\textit{1,2,10}&
		\textit{2:7}& \textit{3,5} & 			 
		\textit{-}&\textit{8:10}\\
		\textbf{3} & \textit{-} &\textit{1:5}&
		\textit{6:10}& \textit{1,2,6:8} & 			
		\textit{-}&\textit{1:10}\\
		\textbf{9} & \textit{-} &\textit{1:3}&
		\textit{5:7}& \textit{-} & 			 
		\textit{1:9}&\textit{10}\\
		\textbf{null} & \textit{3,4} &\textit{5}&
		\textit{10}& \textit{-} & 			 
		\textit{6,7}&\textit{1:3,8,9}\\\\
		\cmidrule(lr){1-7}
		\multicolumn{7}{l}{}
		\begin{minipage}{10cm}
			\textsuperscript{*}\footnotesize{State transitions used to simulate sequence of images with time-varying visual patterns. Numbers in cells are the $\delta$ needed to take a determinate path between $S_t$ and $S_{t+1}$ e.g. `7:10' means $\delta \in [7,10]$, `1:3,5:9' means $\delta \in [1,3]$ or $\delta \in [5,9]$ and so on.}
		\end{minipage}
		\\
		\bottomrule
	\end{tabular}
\end{table}

\begin{longtable}{r P{1cm} P{1cm} P{1cm} P{1cm} P{1cm} P{1cm}}
	\caption{Results on simulated data\textsuperscript{*}}\\*
	\label{tab:results-sim-table}
	\centering
	\begin{tabular}{r P{1cm} P{1cm} P{1cm} P{1cm} P{1cm} P{1cm}}
		\toprule
		\multirow{4}{*}{\textbf{AlexNet}} & \multicolumn{6}{c}{\textbf{Labels}} \\
		\midrule
		& \textbf{0} & \textbf{6} & \textbf{8} & \textbf{3} & \textbf{9}& \textbf{avg.}\\
		\cmidrule(lr){2-7}
		\textbf{PPV} & \textit{$0.5721$} &\textit{$ 0.8303$}&
		\textit{$0.7771$}& \textit{$0.5157$} & 			 
		\textit{$0.8199$} & \textit{$0.7030$}\\
		\textbf{NPV} & \textit{$0.8096$} &\textit{$0.8922$}&
		\textit{$0.9143$}& \textit{$0.9068$} & 	\textit{$0.9410$} & \textit{$0.8927$}\\
		\textbf{F-measure} & \textit{$0.5113$} &\textit{$0.8793$}&
		\textit{$0.8451$}& \textit{$0.4869$} & 			
		\textit{$0.8928$}&\textit{$0.7231$}\\
		\midrule
		\textbf{LSTM} &&&&&\\
		\cmidrule(lr){2-7}
		\textbf{PPV} & \textit{$0.5223$} &\textit{$0.8448$}&
		\textit{$0.7834$}& \textit{$0.4847$} & 			 
		\textit{$0.8314$} & \textit{$0.6933$}\\
		\textbf{NPV} & \textit{$0.8345$} &\textit{$0.8716$}&
		\textit{$0.9050$}& \textit{$0.9147$} & 	\textit{$0.9552$} & \textit{$0.8962$}\\
		\textbf{F-measure} & \textit{$0.5514$} &\textit{$0.8794$}&
		\textit{$0.8445$}& \textit{$0.5028$} & 			
		\textit{$0.9014$}&\textit{$0.7359$}\\
		\midrule
		\textbf{tLSTMv2} &&&&&\\
		\cmidrule(lr){2-7}
		\textbf{PPV} & \textit{${0.9350}$} &\textit{$0.9516$}&
		\textit{$0.9459$}& \textit{$0.7050$} & 			 
		\textit{$0.9838$} & \textit{$0.9043$}\\
		\textbf{NPV} & \textit{$0.9255$} &\textit{$0.9443$}&
		\textit{$0.9517$}& \textit{$0.9413$} & 	\textit{$0.9816$} & \textit{$0.9489$}\\
		\textbf{F-measure} & \textit{$0.8579$} &\textit{$0.9556$}&
		\textit{$0.9476$}& \textit{$0.6836$} & 			
		\textit{$0.9870$}&\textit{$0.8864$}\\
		\midrule
		\textbf{tLSTMv1} &&&&&\\
		\cmidrule(lr){2-7}
		\textbf{PPV} & \textit{$0.9209$} &\textit{${0.9823}$}&
		\textit{${0.9714}$}& \textit{${0.7740}$} & 	\textit{${0.9892}$} & \textit{${0.9276}$}\\
		\textbf{NPV} & \textit{$0.9713$} &\textit{$0.9758$}&
		\textit{$0.9620$}& \textit{$0.9641$} & 	\textit{$0.9898$} & \textit{$0.9726$}\\
		\textbf{F-measure} & \textit{$0.9227$} &\textit{$0.9823$}&
		\textit{$0.9654$}& \textit{$0.7856$} & 			
		\textit{$0.9919$}&\textit{$0.9296$}\\
		\cmidrule(lr){1-7}
		\multicolumn{7}{l}{}
		\begin{minipage}{10cm}
			\textsuperscript{*}\footnotesize{Experimental results obtained on simulated data.}
		\end{minipage}
		\\
		\bottomrule
		
	\end{tabular}
	
\end{longtable}

\end{document}